\pdfoutput=1

\documentclass[11pt]{article}

\usepackage[preprint]{acl}

\usepackage{times}
\usepackage{latexsym}
\usepackage{amsmath}
\usepackage{algorithm}
\usepackage[noend]{algpseudocode}
\usepackage{tabularx}
\usepackage{bm}
\usepackage{amsfonts}
\usepackage{amssymb}
\usepackage{acl}
\usepackage{verbatim}

\usepackage{placeins}
\usepackage{afterpage}

\usepackage[T1]{fontenc}

\usepackage[utf8]{inputenc}

\usepackage{microtype}

\usepackage{inconsolata}

\usepackage{graphicx}
\usepackage{booktabs}
\usepackage{natbib}
\usepackage{makecell}
\usepackage{multirow}
\usepackage{colortbl}
\title{Training LayoutLM from Scratch for Efficient Named-Entity Recognition\\in the Insurance Domain}

\author{Benno Uthayasooriyar\textsuperscript{{\normalfont 1, 2}} \qquad Antoine Ly\textsuperscript{{\normalfont 1}} \qquad Franck Vermet\textsuperscript{{\normalfont 2}} \qquad Caio Corro \textsuperscript{{\normalfont 2}}
\\
\textsuperscript{1}Data Analytics Solutions, SCOR
\quad
\textsuperscript{2}Univ Brest, CNRS, UMR 6205, LMBA
\\
\textsuperscript{3}INSA Rennes, IRISA, Inria, CNRS, Université de Rennes
}

\begin{document}
\maketitle
\begin{abstract}
Generic pre-trained neural networks may struggle to produce good results in specialized domains like finance and insurance.
This is due to a domain mismatch between training data and downstream tasks, as in-domain data are often scarce due to privacy constraints.
In this work, we compare different pre-training strategies for \textsc{LayoutLM}.
We show that using domain-relevant documents improves results on a named-entity recognition (NER) problem using a novel dataset of anonymized insurance-related financial documents called \textsc{Payslips}.
Moreover, we show that we can achieve competitive results using a smaller and faster model.
\end{abstract}

\section{Introduction}
\label{sec:intro}

Modern natural language processing pipelines heavily rely on pre-trained neural networks, primarily language models \cite[][\emph{inter alia}]{schwenk2005deeplm,jozefowicz2016lm,radford2019gpt2} and context-sensitive embeddings \cite[][\emph{inter alia}]{schutze1998wordsense,peters2018elmo,devlin-etal-2019-bert}.
The development of neural architectures based on the attention mechanism \cite{bahdanau2015attention} allows to efficiently pre-train them on GPU using large datasets \cite{vaswani2017attention}: most recent networks can contain several hundreds of billions parameters \cite[\emph{e.g.,}][]{chowdhery2022palm}.

Despite their experimental success, commercial use of pre-trained neural networks can be limited for the following reasons.
Firstly, downstream tasks in information retrieval may require to continuously analyze large amounts of data, which prevents the use of the largest models due to inference time bottleneck.
Secondly, applications in specific fields such as financial, medical or insurance, can forbid the use of API-based models due to privacy concerns.
Thirdly and lastly, authors may at some point decide to not publicly share latest versions of their models, or to change the license to forbid commercial use.\footnote{See for example \textsc{LayoutLMv2} and \textsc{LayoutLMv3}.}
As such, it is increasingly important to ensure replicability and robustness to changes in training data (including for domain transfer) not only for scientific reasons, but also to ensure widespread commercial deployment.

\begin{figure}
    \centering
    \includegraphics[width=\columnwidth]{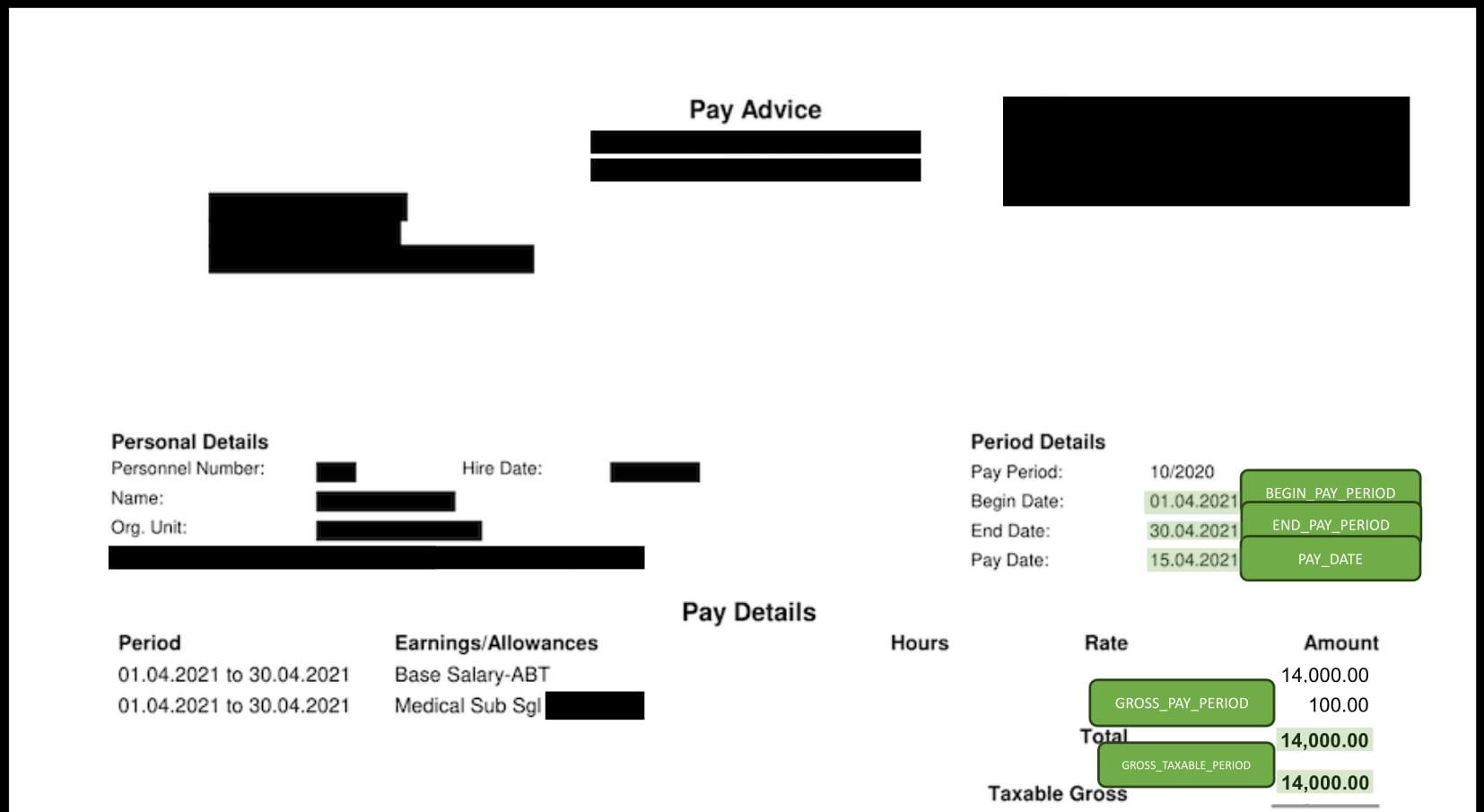}
    \caption{Sample of the newly introduced \textsc{Payslips} dataset for named-entity recognition in the insurance domain.}
    \label{fig:labeled_payslip}
\end{figure}

In this work, we study \textsc{LayoutLM} \cite{xu_layoutlm_2020} for named-entity recognition (NER) on financial documents from the insurance domain.
Our aim is to understand how such a model can be used in a constrained setting: 
Can performance in downstream tasks be improved by pre-training on domain-specific documents, even when the amount of available data is limited? 
Can inference time be improved while maintaining downstream performance?
To address this, we pre-train several models from scratch using a smaller, but more relevant, set of publicly available documents. 

To evaluate these models, we build a novel dataset, \textsc{Payslips}, that contains anonymized insurance pay statements with annotated financial information for NER, detailed in Table \ref{tab:payslips_description}.
Although these documents are private, we have manually anonymized them.
Our experiments show that pre-training on documents that are semantically and structurally similar to those in the downstream task leads to improved performance, even with less training data.
Moreover, if speed of inference is crucial, we show that comparable results can be obtained by using only half the number of layers compared to the original \textsc{LayoutLM} model.

Our contributions can be summarized as follows:
\begin{itemize}
    \item We build and release \textsc{Payslips}, a novel NER dataset of 611 labeled pages of anonymized payslips from the insurance domain;
    \item We pre-train a \textsc{LayoutLM} network using a smaller set of documents \cite[\textsc{DOCILE},][]{simsa_docile_2023};
    \item We evaluate our model on \textsc{Payslips} and show that not only does it achieve better F1 scores, but it also has a lower variance;
    \item We show that a smaller model with half the number of layers maintains performances while improving computational efficiency.
\end{itemize}
Our code and data are publicly available.\footnote{\url{https://github.com/buthaya/payslips}}
\section{Related Work}

\textbf{Contextual embeddings.}
\citet{peters2018elmo} first proposed to pre-train a bidirectionnal LSTM on large corpora to learn context-sensitive word embeddings that can be used to improve results on downstream tasks.
The \textsc{Bert} model \citep{devlin-etal-2019-bert} instead uses a self-attentive network (i.e.\ the encoder part of a transformer) to take full advantage of GPU architectures.
However, \textsc{Bert} cannot be trained using the standard language modeling objective as it is not an autoregressive model.
Instead, the authors proposed a \emph{masked language modeling} objective where the loss aims to increase a reconstruction term on a hidden part of the input.

\textbf{Document analysis.}
For document processing, one must take into account spatial information together with textual content.
\textsc{LayoutLM} \cite{xu_layoutlm_2020} extends \textsc{Bert}'s positional embeddings with spatial positions.
In other words, \textsc{Bert} uses as input embeddings representing the position in the sequence,\footnote{Note that some models uses positional encoding without relying on an embedding table, see for examples \cite[Section 3.5]{vaswani2017attention}} whereas \textsc{LayoutLM} also includes 6-tuples of embeddings describing (discretized) positions and sizes of the boxes containing one or several words.
This allows the self-attentive network to capture spatial information, which is especially useful for documents containing tables and/or processed with optical character recognition.\footnote{OCR's outputs are composed of boxes containing part of the document text.}

\textsc{LayoutLMv2} \cite{xu_layoutlmv2_2022} and \textsc{v3} \cite{huang_layoutlmv3_2022} incorporate more visual information, both as input and in auxiliary training losses.
Moreover, the architecture is modified to integrate relative positional information.
\citet{li_structurallm_2021} introduced richer positional information,
whereas \citet{wang_lilt_2022} focused on language adaptation during the fine-tuning phase.
Contrary to these works, we focus on the original \textsc{LayoutLM} model as we aim for computational efficiency.

\textbf{Efficient encoders.}
The self-attention mechanism has a quadratic-time complexity with respect to the input, which can be slow for long documents.
Several works in document analysis \cite[][\emph{inter alia}]{nguyen-etal-2021-skim-attention,douzon_long-range_2023} have addressed these drawbacks by integrating more computationally efficient types of attention that are better motivated for document processing.
In this work, we instead explore the impact of the number of layers on downstream results.
\section{Payslips Dataset}
\label{sec:payslips}

\begin{figure*}[h]
    \centering    
    \includegraphics[width=0.7\textwidth]{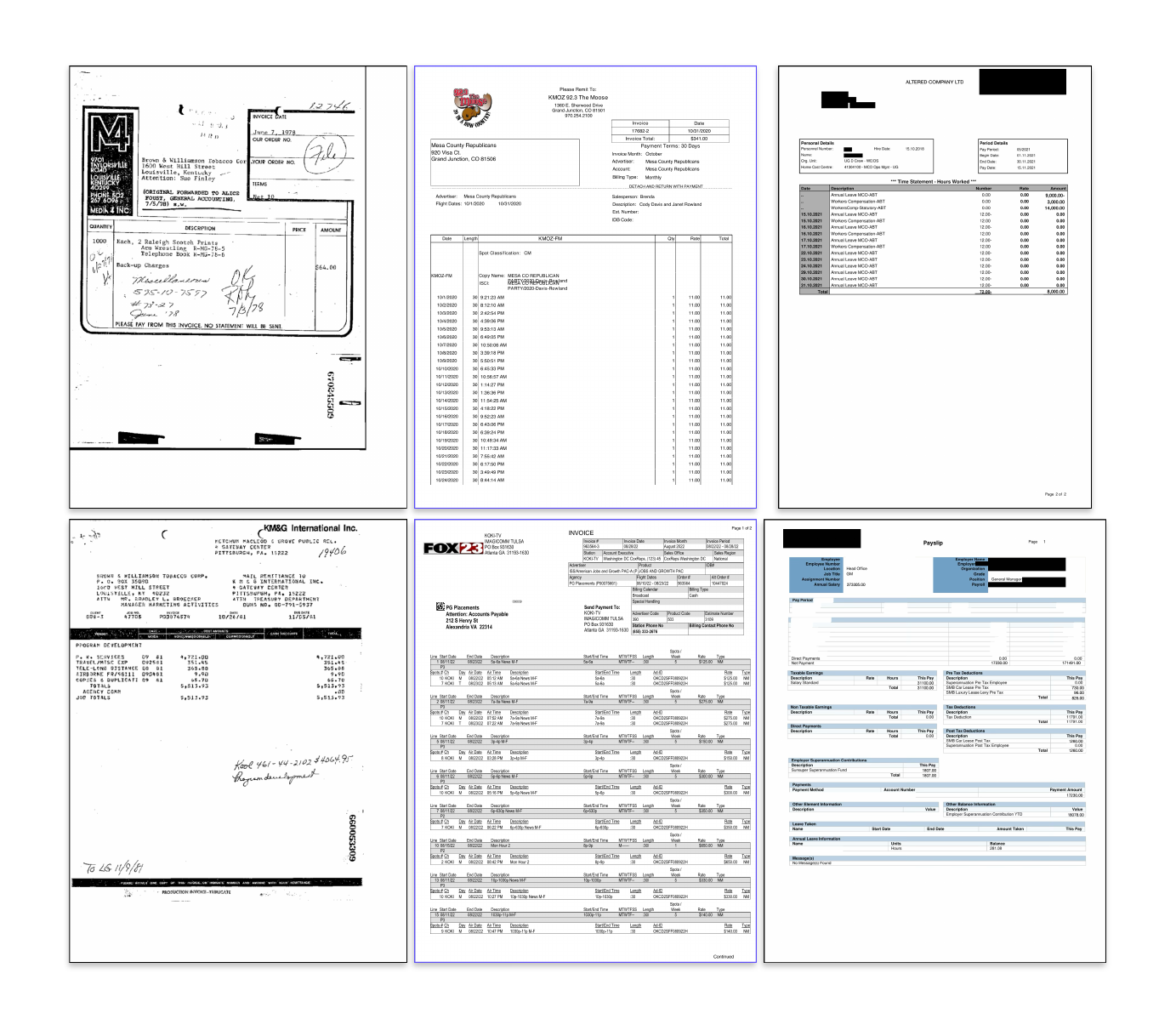}
    \caption{Samples from \textsc{IIT-CDIP} (first column), \textsc{DOCILE} (second column) and \textsc{Payslips} (third column) datasets. Invoices from \textsc{DOCILE} and pay statements from \textsc{Payslips} are closer visually and semantically.}
    \label{fig:sample_data}
\end{figure*}

\begin{table}
\centering\small
\begin{tabular}{@{}l@{\hspace{0.05in}}r@{\hspace{0.05in}}r@{}}
\textbf{Label} & \textbf{Train} & \textbf{Test} \\
\midrule
BEGIN\_PAY\_PERIOD                    & 236   & 85   \\
END\_PAY\_PERIOD                      & 388   & 100  \\
PAY\_DATE                             & 461   & 101  \\
GROSS\_PAY\_PERIOD                    & 481   & 117  \\
GROSS\_TAXABLE\_PERIOD                & 245   & 90   \\
NET\_PAY\_PERIOD                      & 444   & 109  \\
PAYG\_TAX\_PERIOD                     & 499   & 119  \\
PRE\_TAX\_DEDUCTION\_PERIOD           & 278   & 68   \\
POST\_TAX\_DEDUCTION\_PERIOD          & 243   & 67   \\
O                                     & 60,596 & 23,228 \\
\midrule
\textbf{Total}  & 63,871 & 24,084 \\
\end{tabular}
\caption{Label distribution in \textsc{Payslips} dataset (word level).}
\label{tab:payslips_description}
\end{table}

We build a novel dataset containing financial pay statements from the insurance domain which we call \textsc{Payslips}.
This dataset consists of a training set of 485 pages and a test set of 126 pages.

The documents originate from data of disability insurance.
In the event of a work-related accident, this insurance product compensates the insured person during their recovery period.
To determine the indemnity amount, the insurer verifies salary information from each insured person's payslip.
To speed up information processing, it is essential to build tools capable of automatically extracting the useful financial information.
To this end, we worked with insurance professionals and  identified nine specific fields, as detailed in Table~\ref{tab:payslips_description}.
The task is therefore reduced to a standard NER problem, similar to what is done in the \textsc{FUNSD} dataset \cite{jaume2019funsd}.
Unlike datasets such as \textsc{FUNSD} or \textsc{CORD} \cite{park2019cord}, \textsc{Payslips} is notably sparse, with a predominant \textsc{O} class, which poses a challenge for the information extraction task.

\textsc{Payslips} was annotated in-house by people familiar with the documents.
Then, samples have been validated by insurance specialists to ensure annotation quality. 
More details about the annotation process are given in Appendix~\ref{app:payslips}.

For privacy reasons, unnecessary or potentially identifying information was altered or deleted.
Moreover, images are not shared as they are not used by our \textsc{LayoutLM} model, but could give visual cues about the entity emitting the files.

\section{LayoutLM}

\subsection{Neural Architecture}

We use the \textsc{LayoutLM} model \cite{xu_layoutlm_2020}, which is based on the same neural architecture as \textsc{Bert} \cite{devlin-etal-2019-bert}, but where inputs are tailored to represent texts in 2D documents.

Given a document where the content has been divided into text blocks, each individual block is encoded as follows: (1)~words are tokenized;
(2)~each token is represented by an embedding;
(3)~2D positional embeddings are added to word embeddings.
The 2D positional embeddings are 6-tuples representing the coordinates of the block in the page's image, and its height and width, discretized and normalized between 0 and 1000.

The original \textsc{LayoutLM} could also incorporate an image embedding derived from a vision model.
We do not include this input as it slows down the model without significant impact on downstream task results --- sometimes the impact is even negative, see \cite[Table~4]{xu_layoutlm_2020}.

Then, for the self-attentive part, we use the \textsc{Base} model, which consists of 12 self-attentive layers. Each layer contains 12 heads of dimension 768, as originally defined by \citet{devlin-etal-2019-bert}.
Finally, during pre-training, the output contextual embeddings are projected into the vocabulary space using a linear layer to compute output logits.

\subsection{Pre-training}

\textbf{Loss.}
We pre-train the model using a Masked Language Modeling (MLM) loss, where part of the input is replaced by dummy embeddings and a negative log-likelihood loss aims to reconstruct the masked part.
\citet{xu_layoutlm_2020} also experimented with a
Multi-label Document Classification (MDC) loss, a supervised task aiming to classify each page into predefined categories.
Their results show that MDC degrades performances, therefore we do not include this loss during pre-training.

\textbf{Data.}
\textsc{LayoutLM} was pre-trained on the \textsc{IIT-CDIP} dataset \cite{schmidt2002tobacco, IIT_CDIP}, which gathers 11 millions documents from the U.S.\ state lawsuits against the tobacco industry in the 1990s.
The authors show that pre-training on this data improves results on several downstream tasks, including NER on \textsc{SROIE} \cite{SROIE_Huang_2019} and \textsc{FUNSD} \cite{jaume2019funsd}.
Unfortunately, during preliminary experiments we observed that \textsc{LayoutLM} tends to under-perform on our internal data.
We suspect \textsc{IIT-CDIP} documents are too different in form and content from insurance documents (see Figure \ref{fig:sample_data}).
Moreover, adapting information retrieval systems to the insurance domain poses significant challenges due to the sensitivity of the data involved, i.e.\ we cannot train and distribute models based on internal data due to private data protection laws.

We found no existing datasets of pay statements.
However, some relevant invoice datasets are available. \citet{limam_2023_8261508} provides a dataset of generated invoices, and \textsc{RVL-CDIP} \cite{rvl_cdip} includes a subset of invoices from the \textsc{IIT-CDIP} collection.
A more recent and larger dataset, \textsc{DOCILE} \cite{simsa_docile_2023}, offers a better match in terms of layout and semantics with our downstream task dataset, \textsc{Payslips}, see Figure~\ref{fig:sample_data}.
It contains approximately 900k unlabeled invoices sourced from two public repositories.\footnote{\url{https://www.industrydocuments.ucsf.edu/}}\footnote{\url{https://publicfiles.fcc.gov}}
Although it is more than 10 times smaller than \textsc{IIT-CDIP}, our experimental results shows that it is big enough for pre-training \textsc{LayoutLM}.

\textbf{Technical details.}
We pre-train LayoutLM from scratch with the MLM loss on the \textsc{DOCILE} dataset, with similar settings to \citet{xu_layoutlm_2020}.
We use a minibatch size of $80$, and ran the training for 5 epochs with a learning rate of $5\times 10^{-5}$.
We use a cosine scheduler with warmup on $5\%$ of the updates.
Pre-training is done on 8 NVIDIA Tesla V100 16GB GPUs.

\section{Experiments}

We tackle the NER problem using the standard \textsc{BIO}-tagging approach \cite{ramshaw1995bio}, i.e. each token is tagged with either \textsc{O} (not in a mention), \textsc{B-Label} (beginning of a mention) or \textsc{I-Label} (inside of a mention), where \textsc{Label} is any mention label allowed in the dataset.
We can then trivially rebuild the full predicted mentions from the predicted \textsc{BIO} tags.

We fine-tune all models with a batch size of $16$ for $10$ epochs, with a fixed learning rate of $5\times 10^{-5}$.

\subsection{Results}
\label{subsec:experimental_setup}

\begin{table}
\footnotesize
\centering
\begin{tabular}{lcc}
\toprule
 \textbf{Model}  & \makecell{\textbf{F1} \\ \textbf{\textsc{Docile} labeled}} & \makecell{\textbf{F1} \\ \textbf{\textsc{Payslips}}} \\
\midrule
\multicolumn{3}{l}{\textbf{Pre-training on \textsc{IIT-CDIP}}}\\
\midrule
$\textsc{LayoutLM}_{\textrm{BASE}}$     & \boldmath{$58.35\pm1.63$} & \underline{$62.31\pm5.13$}  \\
\midrule
\multicolumn{3}{l}{\textbf{Pre-training on \textsc{Docile}}}\\
\midrule
$\textsc{LayoutLM}_{\textrm{BASE}}$     & \underline{$58.30\pm1.52$} & \boldmath{$64.74\pm2.92$}\\
$\textsc{LayoutLM}_{\textrm{6 layers}}$ & $57.38\pm1.38$ & $61.80\pm3.12$\\
$\textsc{LayoutLM}_{\textrm{2 layers}}$ & $53.89\pm1.03$ & $54.61\pm3.71$ \\
$\textsc{LayoutLM}_{\textrm{1 layer}}$  & $51.12\pm1.53$ & $45.08\pm3.31$ \\
\bottomrule
\end{tabular}
\caption{F1 scores for named-entity recognition using different pre-training and fine-tuning datasets. Results are averaged on 100 runs with different seeds.
}
\label{tab:results}
\end{table}

\begin{table}
    \centering
    \small
    \begin{tabular}{lc}
        \toprule
        \textbf{Model} & \textbf{Inference Time (ms)}  \\
        \midrule
        $\textsc{LayoutLM}_{\textrm{BASE}}$ & $12.10$  \\
        $\textsc{LayoutLM}_{\textrm{6 layers}}$ & $6.15$ \\
        $\textsc{LayoutLM}_{\textrm{2 layers}}$ & $2.42$   \\
        $\textsc{LayoutLM}_{\textrm{1 layers}}$ & $1.73$  \\
        \bottomrule        
    \end{tabular}
    \caption{Inference times per page on the \textsc{Payslips} dataset. Tests were conducted on a machine equipped with a single NVIDIA Tesla V100 32GB GPU.}
    \label{tab:inference_times}
\end{table}

We compare the original \textsc{LayoutLM} pre-trained on \textsc{IIT-CDIP} with our \textsc{LayoutLM} pre-trained (from scratch) on \textsc{DOCILE} on two NER datasets:
(1) The subset of the \textsc{DOCILE} dataset which is labeled\footnote{As the annotation of the test set are not available online, we performed evaluation on the validation set.} --- it contains 6759 and 635 document pages for training and testing, respectively;
(2) Our novel \textsc{Payslips} dataset --- statistics are reported in Section~\ref{sec:payslips}.
We fine-tune similarly for both datasets.

We report labeled F1-score averaged on $100$ fine-tuning runs in Table~\ref{tab:results}.
Precision and recall are reported in Appendix~\ref{app:results}.
The \textsc{Base} model (using the full 12 layers) produces similar results on \textsc{DOCILE} no matter if pre-training on \textsc{IIT-CDIP} or \textsc{DOCILE}.
However, on our internal \textsc{Payslips} datasets, our model pre-trained on \textsc{DOCILE} outperforms the original one.
Moreover, we observe that our pre-trained model exhibits a way lower variance between fine-tuning runs.

To cope with the high and continuous flow of documents, an insurer might require a faster model. Therefore, we also experimented using a smaller number of self-attentive layers, see Table~\ref{tab:results}.
Inference times per model are reported in Table~\ref{tab:inference_times}.
On \textsc{Payslips}, when pre-training on \textsc{DOCILE} using only 6 layers, we achieve comparable scores to the off-the-shelf \textsc{LayoutLM} model, while dividing the inference time by almost 2.

\subsection{Statistical Significance}\label{subsec:statistical_significance}


Domain-specific datasets are often of small sizes, so comparing F1-scores may lead to wrong conclusion if they are not statistically significant.
We follow the original Message Understanding Conference \cite[MUC,][]{chinchor-1992-statistical, chinchor-etal-1993-evaluating} and rely on the approximate randomization method \cite{noreen_computer-intensive_1989}, which does not require assumptions on the data distribution.
For this test, the null hypothesis is
``\textit{The proposed system and the baseline system do not differ in F1}''.
The difference is computed in term of absolute F1 difference over many random data splits.
Pseudo-code is given in Appendix~\ref{app:signif}.

In our case, we compare the $\textsc{LayoutLM}$ pre-trained on \textsc{IIT-CDIP} to the one pre-trained on \textsc{DOCILE}, both being fine-tuned on \textsc{Payslips}. 
As we did 100 fine-tunings, we took two models with a F1-score difference below $1.00$ for the test.
The obtained significance value, $0.0019$, is lower than $0.01$ and thus considered highly significant, according to \citet[Figure~3]{chinchor-1992-statistical}.

\section{Conclusion}

In this work, we pre-train from scratch a \textsc{LayoutLM} model using the \textsc{DOCILE} dataset.
Importantly, we show that our model obtain better results on a novel domain-specific NER dataset.
This shows that it is still possible to develop fast and state-of-the-art in-house models that allow commercial usage.

We also release our novel \textsc{Payslips} dataset that can be used to challenge document processing models in financial domains.
\section{Limitations}

The novel \textsc{Payslips} dataset is of small size compared to many standard benchmarks.
Unfortunately, specialized domains like insurance not only induce expensive annotation costs,
but it is also difficult to obtain authorization to publicly release the data.
This issue is also common in other domains like biomedical NLP.
Another issue is that \textsc{Payslips} is highly specialized, so interest may be limited.

Experimental results highlight that NLP models may not be useful for production yet, as the F1 scores are below 65.
\section*{Acknowledgement}

We thank the anonymous reviewers for their comments and suggestions.
This work was performed using HPC resources from GENCI-IDRIS (Grant 2024-AD011015001).

\bibliography{custom}

\begin{thebibliography}{28}
\providecommand{\natexlab}[1]{#1}

\bibitem[{Bahdanau et~al.(2015)Bahdanau, Cho, and
  Bengio}]{bahdanau2015attention}
Dzmitry Bahdanau, Kyunghyun Cho, and Yoshua Bengio. 2015.
\newblock \href {http://arxiv.org/abs/1409.0473} {Neural machine translation by
  jointly learning to align and translate}.
\newblock In \emph{ICLR}.

\bibitem[{Chinchor(1992)}]{chinchor-1992-statistical}
Nancy Chinchor. 1992.
\newblock \href {https://aclanthology.org/M92-1003} {The statistical
  significance of the {MUC}-4 results}.
\newblock In \emph{{F}ourth {M}essage {U}nderstanding {C}onference ({MUC}-4):
  Proceedings of a Conference Held in {M}c{L}ean, {V}irginia, {J}une 16-18,
  1992}.

\bibitem[{Chinchor et~al.(1993)Chinchor, Hirschman, and
  Lewis}]{chinchor-etal-1993-evaluating}
Nancy Chinchor, Lynette Hirschman, and David~D. Lewis. 1993.
\newblock \href {https://aclanthology.org/J93-3001} {Evaluating message
  understanding systems: An analysis of the third {M}essage {U}nderstanding
  {C}onference ({MUC}-3)}.
\newblock \emph{Computational Linguistics}, 19(3):409--450.

\bibitem[{Chowdhery et~al.(2023)Chowdhery, Narang, Devlin, Bosma, Mishra,
  Roberts, Barham, Chung, Sutton, Gehrmann, Schuh, Shi, Tsvyashchenko, Maynez,
  Rao, Barnes, Tay, Shazeer, Prabhakaran, Reif, Du, Hutchinson, Pope, Bradbury,
  Austin, Isard, Gur-Ari, Yin, Duke, Levskaya, Ghemawat, Dev, Michalewski,
  Garcia, Misra, Robinson, Fedus, Zhou, Ippolito, Luan, Lim, Zoph, Spiridonov,
  Sepassi, Dohan, Agrawal, Omernick, Dai, Pillai, Pellat, Lewkowycz, Moreira,
  Child, Polozov, Lee, Zhou, Wang, Saeta, Diaz, Firat, Catasta, Wei,
  Meier-Hellstern, Eck, Dean, Petrov, and Fiedel}]{chowdhery2022palm}
Aakanksha Chowdhery, Sharan Narang, Jacob Devlin, Maarten Bosma, Gaurav Mishra,
  Adam Roberts, Paul Barham, Hyung~Won Chung, Charles Sutton, Sebastian
  Gehrmann, Parker Schuh, Kensen Shi, Sasha Tsvyashchenko, Joshua Maynez,
  Abhishek Rao, Parker Barnes, Yi~Tay, Noam Shazeer, Vinodkumar Prabhakaran,
  Emily Reif, Nan Du, Ben Hutchinson, Reiner Pope, James Bradbury, Jacob
  Austin, Michael Isard, Guy Gur-Ari, Pengcheng Yin, Toju Duke, Anselm
  Levskaya, Sanjay Ghemawat, Sunipa Dev, Henryk Michalewski, Xavier Garcia,
  Vedant Misra, Kevin Robinson, Liam Fedus, Denny Zhou, Daphne Ippolito, David
  Luan, Hyeontaek Lim, Barret Zoph, Alexander Spiridonov, Ryan Sepassi, David
  Dohan, Shivani Agrawal, Mark Omernick, Andrew~M. Dai,
  Thanumalayan~Sankaranarayana Pillai, Marie Pellat, Aitor Lewkowycz, Erica
  Moreira, Rewon Child, Oleksandr Polozov, Katherine Lee, Zongwei Zhou, Xuezhi
  Wang, Brennan Saeta, Mark Diaz, Orhan Firat, Michele Catasta, Jason Wei,
  Kathy Meier-Hellstern, Douglas Eck, Jeff Dean, Slav Petrov, and Noah Fiedel.
  2023.
\newblock \href {http://jmlr.org/papers/v24/22-1144.html} {Palm: Scaling
  language modeling with pathways}.
\newblock \emph{Journal of Machine Learning Research}, 24(240):1--113.

\bibitem[{Devlin et~al.(2019)Devlin, Chang, Lee, and
  Toutanova}]{devlin-etal-2019-bert}
Jacob Devlin, Ming-Wei Chang, Kenton Lee, and Kristina Toutanova. 2019.
\newblock \href {https://doi.org/10.18653/v1/N19-1423} {{BERT}: Pre-training of
  deep bidirectional transformers for language understanding}.
\newblock In \emph{Proceedings of the 2019 Conference of the North {A}merican
  Chapter of the Association for Computational Linguistics: Human Language
  Technologies, Volume 1 (Long and Short Papers)}, pages 4171--4186,
  Minneapolis, Minnesota. Association for Computational Linguistics.

\bibitem[{Douzon et~al.(2023)Douzon, Duffner, Garcia, and
  Espinas}]{douzon_long-range_2023}
Thibault Douzon, Stefan Duffner, Christophe Garcia, and J{\'e}r{\'e}my Espinas.
  2023.
\newblock \href
  {https://link.springer.com/chapter/10.1007/978-3-031-41501-2_4#citeas}
  {Long-range transformer architectures for document understanding}.
\newblock In \emph{Document Analysis and Recognition -- ICDAR 2023 Workshops},
  pages 47--64, Cham. Springer Nature Switzerland.

\bibitem[{Harley et~al.(2015)Harley, Ufkes, and Derpanis}]{rvl_cdip}
Adam~W Harley, Alex Ufkes, and Konstantinos~G Derpanis. 2015.
\newblock \href {https://adamharley.com/icdar15/<} {Evaluation of deep
  convolutional nets for document image classification and retrieval}.
\newblock In \emph{International Conference on Document Analysis and
  Recognition ({ICDAR})}.

\bibitem[{Huang et~al.(2022)Huang, Lv, Cui, Lu, and
  Wei}]{huang_layoutlmv3_2022}
Yupan Huang, Tengchao Lv, Lei Cui, Yutong Lu, and Furu Wei. 2022.
\newblock \href {https://doi.org/10.1145/3503161.3548112} {Layoutlmv3:
  Pre-training for document ai with unified text and image masking}.
\newblock In \emph{Proceedings of the 30th ACM International Conference on
  Multimedia}, MM '22, page 4083–4091, New York, NY, USA. Association for
  Computing Machinery.

\bibitem[{Huang et~al.(2019)Huang, Chen, He, Bai, Karatzas, Lu, and
  Jawahar}]{SROIE_Huang_2019}
Zheng Huang, Kai Chen, Jianhua He, Xiang Bai, Dimosthenis Karatzas, Shijian Lu,
  and C.~V. Jawahar. 2019.
\newblock \href {https://doi.org/10.1109/ICDAR.2019.00244} {Icdar2019
  competition on scanned receipt ocr and information extraction}.
\newblock In \emph{2019 International Conference on Document Analysis and
  Recognition (ICDAR)}, pages 1516--1520.

\bibitem[{Jaume et~al.(2019)Jaume, Kemal~Ekenel, and Thiran}]{jaume2019funsd}
Guillaume Jaume, Hazim Kemal~Ekenel, and Jean-Philippe Thiran. 2019.
\newblock \href {https://doi.org/10.1109/ICDARW.2019.10029} {Funsd: A dataset
  for form understanding in noisy scanned documents}.
\newblock In \emph{2019 International Conference on Document Analysis and
  Recognition Workshops (ICDARW)}, volume~2, pages 1--6.

\bibitem[{Jozefowicz et~al.(2016)Jozefowicz, Vinyals, Schuster, Shazeer, and
  Wu}]{jozefowicz2016lm}
Rafal Jozefowicz, Oriol Vinyals, Mike Schuster, Noam Shazeer, and Yonghui Wu.
  2016.
\newblock \href {https://arxiv.org/abs/1602.02410} {Exploring the limits of
  language modeling}.
\newblock \emph{Preprint}, arXiv:1602.02410.

\bibitem[{Lewis et~al.(2006)Lewis, Agam, Argamon, Frieder, Grossman, and
  Heard}]{IIT_CDIP}
D.~Lewis, G.~Agam, S.~Argamon, O.~Frieder, D.~Grossman, and J.~Heard. 2006.
\newblock \href {https://doi.org/10.1145/1148170.1148307} {Building a test
  collection for complex document information processing}.
\newblock In \emph{Proceedings of the 29th Annual International ACM SIGIR
  Conference on Research and Development in Information Retrieval}, SIGIR '06,
  page 665–666, New York, NY, USA. Association for Computing Machinery.

\bibitem[{Li et~al.(2021)Li, Bi, Yan, Wang, Huang, Huang, and
  Si}]{li_structurallm_2021}
Chenliang Li, Bin Bi, Ming Yan, Wei Wang, Songfang Huang, Fei Huang, and Luo
  Si. 2021.
\newblock \href {https://doi.org/10.18653/v1/2021.acl-long.493}
  {{S}tructural{LM}: Structural pre-training for form understanding}.
\newblock In \emph{Proceedings of the 59th Annual Meeting of the Association
  for Computational Linguistics and the 11th International Joint Conference on
  Natural Language Processing (Volume 1: Long Papers)}, pages 6309--6318,
  Online. Association for Computational Linguistics.

\bibitem[{Limam et~al.(2023)Limam, Dhiaf, and Kessentini}]{limam_2023_8261508}
Mahmoud Limam, Marwa Dhiaf, and Yousri Kessentini. 2023.
\newblock \href {https://doi.org/10.5281/zenodo.8261508} {Fatura dataset}.

\bibitem[{Nguyen et~al.(2021)Nguyen, Scialom, Staiano, and
  Piwowarski}]{nguyen-etal-2021-skim-attention}
Laura Nguyen, Thomas Scialom, Jacopo Staiano, and Benjamin Piwowarski. 2021.
\newblock \href {https://doi.org/10.18653/v1/2021.findings-emnlp.207}
  {Skim-attention: Learning to focus via document layout}.
\newblock In \emph{Findings of the Association for Computational Linguistics:
  EMNLP 2021}, pages 2413--2427, Punta Cana, Dominican Republic. Association
  for Computational Linguistics.

\bibitem[{Noreen(1989)}]{noreen_computer-intensive_1989}
E.W. Noreen. 1989.
\newblock \href {https://books.google.fr/books?id=kinvAAAAMAAJ}
  {\emph{Computer-{Intensive} {Methods} for {Testing} {Hypotheses}: {An}
  {Introduction}}}.
\newblock Wiley.

\bibitem[{Park et~al.(2019)Park, Shin, Lee, Lee, Surh, Seo, and
  Lee}]{park2019cord}
Seunghyun Park, Seung Shin, Bado Lee, Junyeop Lee, Jaeheung Surh, Minjoon Seo,
  and Hwalsuk Lee. 2019.
\newblock Cord: A consolidated receipt dataset for post-ocr parsing.

\bibitem[{Peters et~al.(2018)Peters, Neumann, Iyyer, Gardner, Clark, Lee, and
  Zettlemoyer}]{peters2018elmo}
Matthew~E. Peters, Mark Neumann, Mohit Iyyer, Matt Gardner, Christopher Clark,
  Kenton Lee, and Luke Zettlemoyer. 2018.
\newblock \href {https://doi.org/10.18653/v1/N18-1202} {Deep contextualized
  word representations}.
\newblock In \emph{Proceedings of the 2018 Conference of the North {A}merican
  Chapter of the Association for Computational Linguistics: Human Language
  Technologies, Volume 1 (Long Papers)}, pages 2227--2237, New Orleans,
  Louisiana. Association for Computational Linguistics.

\bibitem[{Radford et~al.(2019)Radford, Wu, Child, Luan, Amodei, and
  Sutskever}]{radford2019gpt2}
Alec Radford, Jeff Wu, Rewon Child, David Luan, Dario Amodei, and Ilya
  Sutskever. 2019.
\newblock Language models are unsupervised multitask learners.

\bibitem[{Ramshaw and Marcus(1995)}]{ramshaw1995bio}
Lance Ramshaw and Mitch Marcus. 1995.
\newblock \href {https://aclanthology.org/W95-0107} {Text chunking using
  transformation-based learning}.
\newblock In \emph{Third Workshop on Very Large Corpora}.

\bibitem[{Schmidt et~al.(2002)Schmidt, Butter, and Rider}]{schmidt2002tobacco}
Heidi Schmidt, Karen Butter, and Cynthia Rider. 2002.
\newblock \href {https://www.dlib.org/dlib/september02/schmidt/09schmidt.html}
  {Building digital tobacco industry document libraries at the university of
  california, san francisco library/center for knowledge management}.
\newblock \emph{D-Lib Magazine}, 8(9):1082--9873.

\bibitem[{Sch{\"u}tze(1998)}]{schutze1998wordsense}
Hinrich Sch{\"u}tze. 1998.
\newblock \href {https://aclanthology.org/J98-1004} {Automatic word sense
  discrimination}.
\newblock \emph{Computational Linguistics}, 24(1):97--123.

\bibitem[{Schwenk and Gauvain(2005)}]{schwenk2005deeplm}
Holger Schwenk and Jean-Luc Gauvain. 2005.
\newblock \href {https://aclanthology.org/H05-1026} {Training neural network
  language models on very large corpora}.
\newblock In \emph{Proceedings of Human Language Technology Conference and
  Conference on Empirical Methods in Natural Language Processing}, pages
  201--208, Vancouver, British Columbia, Canada. Association for Computational
  Linguistics.

\bibitem[{Vaswani et~al.(2017)Vaswani, Shazeer, Parmar, Uszkoreit, Jones,
  Gomez, Kaiser, and Polosukhin}]{vaswani2017attention}
Ashish Vaswani, Noam Shazeer, Niki Parmar, Jakob Uszkoreit, Llion Jones,
  Aidan~N Gomez, \L~ukasz Kaiser, and Illia Polosukhin. 2017.
\newblock \href
  {https://proceedings.neurips.cc/paper_files/paper/2017/file/3f5ee243547dee91fbd053c1c4a845aa-Paper.pdf}
  {Attention is all you need}.
\newblock In \emph{Advances in Neural Information Processing Systems},
  volume~30. Curran Associates, Inc.

\bibitem[{Wang et~al.(2022)Wang, Jin, and Ding}]{wang_lilt_2022}
Jiapeng Wang, Lianwen Jin, and Kai Ding. 2022.
\newblock \href {https://doi.org/10.18653/v1/2022.acl-long.534} {{L}i{LT}: A
  simple yet effective language-independent layout transformer for structured
  document understanding}.
\newblock In \emph{Proceedings of the 60th Annual Meeting of the Association
  for Computational Linguistics (Volume 1: Long Papers)}, pages 7747--7757,
  Dublin, Ireland. Association for Computational Linguistics.

\bibitem[{Xu et~al.(2021)Xu, Xu, Lv, Cui, Wei, Wang, Lu, Florencio, Zhang, Che,
  Zhang, and Zhou}]{xu_layoutlmv2_2022}
Yang Xu, Yiheng Xu, Tengchao Lv, Lei Cui, Furu Wei, Guoxin Wang, Yijuan Lu,
  Dinei Florencio, Cha Zhang, Wanxiang Che, Min Zhang, and Lidong Zhou. 2021.
\newblock \href {https://doi.org/10.18653/v1/2021.acl-long.201}
  {{L}ayout{LM}v2: Multi-modal pre-training for visually-rich document
  understanding}.
\newblock In \emph{Proceedings of the 59th Annual Meeting of the Association
  for Computational Linguistics and the 11th International Joint Conference on
  Natural Language Processing (Volume 1: Long Papers)}, pages 2579--2591,
  Online. Association for Computational Linguistics.

\bibitem[{Xu et~al.(2020)Xu, Li, Cui, Huang, Wei, and Zhou}]{xu_layoutlm_2020}
Yiheng Xu, Minghao Li, Lei Cui, Shaohan Huang, Furu Wei, and Ming Zhou. 2020.
\newblock \href {https://doi.org/10.1145/3394486.3403172} {Layoutlm:
  Pre-training of text and layout for document image understanding}.
\newblock In \emph{Proceedings of the 26th ACM SIGKDD International Conference
  on Knowledge Discovery \& Data Mining}, KDD '20, page 1192–1200, New York,
  NY, USA. Association for Computing Machinery.

\bibitem[{Šimsa et~al.(2023)Šimsa, Šulc, Uřičář, Patel, Hamdi, Kocián,
  Skalický, Matas, Doucet, Coustaty, and Karatzas}]{simsa_docile_2023}
Štěpán Šimsa, Milan Šulc, Michal Uřičář, Yash Patel, Ahmed Hamdi,
  Matěj Kocián, Matyáš Skalický, Jiří Matas, Antoine Doucet, Mickaël
  Coustaty, and Dimosthenis Karatzas. 2023.
\newblock \href
  {https://link.springer.com/chapter/10.1007/978-3-031-41679-8_9#citeas}
  {{DocILE} {Benchmark} for {Document} {Information} {Localization} and
  {Extraction}}.
\newblock In \emph{Document {Analysis} and {Recognition} - {ICDAR} 2023}, pages
  147--166, Cham. Springer Nature Switzerland.

\end{thebibliography}

\appendix
\clearpage

\section{Statistical Significance}
\label{app:signif}

In the context of working with small test sets, it is important to validate that differences in experimental results are not attributable to randomness. 
To achieve this, (1) we run 100 times each fine-tuning experiment, using different random seeds for both data shuffling and initialization of the linear layer, and (2) we conduct statistical significance testing. 

We follow the same procedure as the Message Understanding Conference \citep{chinchor-1992-statistical, chinchor-etal-1993-evaluating} and rely on approximate randomization testing.
This test is performed on a test set using two systems, A and B. 
For $9999$ iterations, the test compares: (1) the difference in average F1-score between A and B on the test set with (2) the difference in average F1-score between two shuffled sets, each containing a mix of the F1-scores of A and B on the test set. 
The significance level is then computed as the percentage of iterations in which the difference in F1-score of the shuffled sets exceeds the actual difference in F1-score between A and B. 
The entire pseudo-code for this test is given in Algorithm~\ref{appendix:approximate_randomization}.

\section{Precision and Recall}
\label{app:results}

In addition to the F1-scores presented in Table~\ref{tab:results}, we provide a detailed precision and recall metrics in Table~\ref{tab:results_appendix}. 
We observe that on \textsc{Payslips}, the gain is mainly due to an increase in precision when pre-training on \textsc{DOCILE}.
It is also interesting to note that when going from 12 to only 6 layers, the drop in performance is, again, due to a drop in precision.



\begin{algorithm}[h!]
\caption{Approximate Randomization testing}
\label{appendix:approximate_randomization}
\begin{algorithmic}[1]
\Function {AR}{$f_{\text{baseline}}$, $f_{\text{proposed}}$, $\mathbf{x}_\text{test}$}
    \State \textbf{Input:} $f_{\text{baseline}}$, $f_{\text{proposed}}$ : the models to compare and $\mathbf{x}_\text{test}$ the test set of size $\mathbf{N}$
    \State \textbf{Output:} $\alpha$ the significance value
    \State $\mathbf{y}_{\text{baseline}} \gets \mathbf{N}$ predictions of the baseline model.
    \State $\mathbf{y}_{\text{proposed}} \gets \mathbf{N}$ predictions of the proposed model.
    \State Compute the mean F1-score for each set of predictions: $\overline{\mathbf{F}_{\text{baseline}}}$ 
    and $\overline{\mathbf{F}_{\text{proposed}}}$
    \State $\Delta F1_{\text{original}} = \left|\overline{\mathbf{F}_{\text{proposed}}} - \overline{\mathbf{F}_{\text{baseline}}}\right|$ 
\State $n_{ge} \gets 0$ \Comment{Counter}
\For{$i \gets 1$ to $9999$}
    \State $\mathbf{y} \gets \mathbf{y}_{\text{baseline}}  \cup \mathbf{y}_{\text{proposed}}$
    \State Shuffle $\mathbf{y}$ 
    \State Split $\mathbf{y}$ into two subsets $\mathbf{y}_A$ and $\mathbf{y}_B$, each of the same size.
    \State Compute the mean F1-score for each shuffled subset: $\overline{\mathbf{F}_A}$ and $\overline{\mathbf{F}_B}$
    \State $\Delta F1_{\text{shuffled}} = \left| \overline{\mathbf{F}_A} - \overline{\mathbf{F}_B} \right|$
    \If{$\Delta F1_{\text{shuffled}} \geq \Delta F1_{\text{original}}$}
        \State $n_{ge} \gets n_{ge} + 1$ \Comment{Increment counter}
    \EndIf
\EndFor
\Return $\alpha = \frac{n_{ge} + 1}{9999 + 1}$
\EndFunction
\Comment{Significance level}
\end{algorithmic}
\end{algorithm}

\begin{table*}
\centering
\begin{tabular}{lcccc}
\toprule
 \textbf{Model} & \textbf{Pre-training dataset}  & \textbf{Precision} & \textbf{Recall} & \textbf{F1} \\
\midrule
\multicolumn{5}{l}{\textbf{Fine-tuned on} \textsc{DOCILE} \textbf{labeled}}\\
\midrule
$\textsc{LayoutLM}_{\textsc{BASE}}$ & \textsc{IIT-CDIP} & \boldmath{$57.79$} & $55.25$ &\boldmath{$58.35\pm1.63$}  \\
$\textsc{LayoutLM}_{\textsc{BASE}}$ & \textsc{DOCILE}  & \underline{$57.22$} & \boldmath{$59.45$} & \underline{$58.30\pm1.52$} \\
$\textsc{LayoutLM}_{\textsc{6 layers}}$ & \textsc{DOCILE}  &$ 56.59$ & \underline{$58.20$} & $57.38\pm1.38$\\
$\textsc{LayoutLM}_{\textsc{2 layers}}$ & \textsc{DOCILE} & $52.65$ & $55.25$ & $53.89\pm1.03$ \\
$\textsc{LayoutLM}_{\textsc{1 layer}}$ & \textsc{DOCILE} & $49.71$ & $52.68$ & $51.12\pm1.53$ \\
\midrule
\multicolumn{5}{l}{\textbf{Fine-tuned on} \textsc{Payslips}}\\
\midrule
$\textsc{LayoutLM}_{\textsc{BASE}}$ & \textsc{IIT-CDIP} & \underline{$65.70$} & \boldmath{$59.80$} & \underline{$62.31\pm5.13$} \\
$\textsc{LayoutLM}_{\textsc{BASE}}$ & \textsc{DOCILE}  & \boldmath{$71.47$} &$ 59.53$ & \boldmath{$64.74\pm2.92$} \\
$\textsc{LayoutLM}_{\textsc{6 layers}}$ & \textsc{DOCILE}  & $64.59$ & \underline{$59.62$} & $61.80\pm3.12$ \\
$\textsc{LayoutLM}_{\textsc{2 layers}}$ & \textsc{DOCILE} & $61.80$ & $49.66$ &$ 54.61\pm3.71$ \\
$\textsc{LayoutLM}_{\textsc{1 layer}}$  & \textsc{DOCILE} & $51.66$ & $40.29$ & $45.08\pm3.31$ \\
\bottomrule
\end{tabular}
\caption{Precision, Recall, and F1 scores for named-entity recognition using different pre-training and fine-tuning datasets. Results are averaged on 100 runs with different seeds.}
\label{tab:results_appendix}
\end{table*}

\section{\textsc{Payslips} construction details}
\label{app:payslips}

The \textsc{Payslips} dataset was obtained to automate the financial assessment at the claims and underwriting stages of a disability product. 
Accelerating this process allows underwriters and claims specialists to focus on less menial tasks while reducing the response time for a new policy or the payment of a claim. 
The underwriting specialists provided the Data Science team with an anonymous version of 611 pay statements. 
These documents were free of non-relevant Personal Identifiable Information (PII) such as names, addresses, ID numbers, and banking information. 
The raw data was then processed through an in-house OCR solution to obtain the text and layout of each page at the word level. 
An extensive annotation procedure was then initiated, during which several Data Scientists followed rules defined with the underwriters regarding the entities to extract. 
As such, only the amounts for the concerned period were annotated, as opposed to the year-to-date (YTD) amounts.
Once the annotation procedure was completed, fine-tuning could be done on this data. 
The results presented in this paper are based on this version of the dataset. 
However, after discussions with SCOR's legal department, we could not share this version of the dataset as it still contained identifiable information about the company issuing the payments and the insured persons. 
To create a shareable version, we had to manually alter several amounts and the remaining sensitive information. 
The amounts were altered while ensuring the consistency and logical relationships between them, to preserve the coherence of the task.

\end{document}